\relax
\documentclass[letterpaper]{article}
\usepackage{aaai17}
\usepackage{times}
\usepackage{helvet}
\usepackage{courier}
\usepackage{graphicx} %additional  TODO: Use this to insert figures
\usepackage[hyphens]{url}
\usepackage{csquotes} %additional
\graphicspath{ {images/} }

\begin{document}

\title{Automated Hate Speech Detection and the Problem of Offensive Language\thanks{This is a preprint of a short paper accepted at ICWSM 2017. Please cite that version instead.}}
\author{Thomas Davidson,\textsuperscript{\textnormal{1}} Dana Warmsley,\textsuperscript{\textnormal{2}} Michael Macy,\textsuperscript{\textnormal{1,3}} Ingmar Weber\textsuperscript{\textnormal{4}}\\
\textsuperscript{1}Department of Sociology, Cornell University, Ithaca, NY, USA \\
\textsuperscript{2}Department of Applied Mathematics, Cornell University, Ithaca, NY, USA \\
\textsuperscript{3}Department of Information Science, Cornell University, Ithaca, NY, USA \\
\textsuperscript{4}Qatar Computing Research Institute, HBKU, Doha, Qatar \\
\{trd54, dw457, mwmacy\}@cornell.edu, iweber@hbku.edu.qa }

\maketitle
\begin{abstract}
A key challenge for automatic hate-speech detection on social media is the separation of hate speech from other instances of offensive language. Lexical detection methods tend to have low precision because they classify all messages containing particular terms as hate speech and previous work using supervised learning has failed to distinguish between the two categories. We used a crowd-sourced hate speech lexicon to collect tweets containing hate speech keywords. We use crowd-sourcing to label a sample of these tweets into three categories: those containing hate speech, only offensive language, and those with neither. We train a multi-class classifier to distinguish between these different categories. Close analysis of the predictions and the errors shows when we can reliably separate hate speech from other offensive language and when this differentiation is more difficult. We find that racist and homophobic tweets are more likely to be classified as hate speech but that sexist tweets are generally classified as offensive. Tweets without explicit hate keywords are also more difficult to classify.
\end{abstract}

\section{Introduction}
What constitutes hate speech and when does it differ from offensive language? No formal definition exists but there is a consensus that it is speech that targets disadvantaged social groups in a manner that is potentially harmful to them \cite{jacobs2000hate,walker1994hate}. In the United States, hate speech is protected under the free speech provisions of the First Amendment, but it has been extensively debated in the legal sphere and with regards to speech codes on college campuses. In many countries, including the United Kingdom, Canada, and France, there are laws prohibiting hate speech, which tends to be defined as speech that targets minority groups in a way that could promote violence or social disorder. People convicted of using hate speech can often face large fines and even imprisonment. These laws extend to the internet and social media, leading many sites to create their own provisions against hate speech. Both Facebook and Twitter have responded to criticism for not doing enough to prevent hate speech on their sites by instituting policies to prohibit the use of their platforms for attacks on people based on characteristics like race, ethnicity, gender, and sexual orientation, or threats of violence towards others.\footnote{Facebook's policy can be found here: \url{www.facebook.com/communitystandards#hate-speech}. Twitter's policy can be found here: \url{support.twitter.com/articles/20175050}.}

Drawing upon these definitions, we define hate speech as \textit{language that is used to expresses hatred towards a targeted group or is intended to be derogatory, to humiliate, or to insult the members of the group}. In extreme cases this may also be language that threatens or incites violence, but limiting our definition only to such cases would exclude a large proportion of hate speech. Importantly, our definition does not include all instances of offensive language because people often use terms that are highly offensive to certain groups but in a qualitatively different manner. For example some African Americans often use the term \textit{n*gga}\footnote{Where present, the ``*'' has been inserted by us and was not part of the original text. All tweets quoted have been modified slightly to protect user's identities while retaining their original meaning.} in everyday language online \cite{Warner:2012:DHS:2390374.2390377}, people use terms like \textit{h*e} and \textit{b*tch} when quoting rap lyrics, and teenagers use homophobic slurs like \textit{f*g} as they play video games. Such language is prevalent on social media \cite{Wang:2014:CET:2531602.2531734}, making this boundary condition crucial for any usable hate speech detection system .

Previous work on hate speech detection has identified this problem but many studies still tend to conflate hate speech and offensive language. In this paper we label tweets into three categories: hate speech, offensive language, or neither. We train a model to differentiate between these categories and then analyze the results in order to better understand how we can distinguish between them. Our results show that fine-grained labels can help in the task of hate speech detection and highlights some of the key challenges to accurate classification. We conclude that future work must better account for context and the heterogeneity in hate speech usage. 

\section{Related Work}
Bag-of-words approaches tend to have high recall but lead to high rates of false positives since the presence of offensive words can lead to the misclassification of tweets as hate speech \cite{DBLP:conf/aaai/KwokW13,burnap2015cyber}. Focusing on anti-black racism, \citeauthor{DBLP:conf/aaai/KwokW13} find that 86\% of the time the reason a tweet was categorized as racist was because it contained offensive words. Given the relatively high prevalence of offensive language and \enquote{curse words} on social media this makes hate speech detection particularly challenging \cite{Wang:2014:CET:2531602.2531734}. The difference between hate speech and other offensive language is often based upon subtle linguistic distinctions, for example tweets containing the word \textit{n*gger} are more likely to be labeled as hate speech than \textit{n*gga} \cite{DBLP:conf/aaai/KwokW13}. Many can be ambiguous, for example the word \textit{gay} can be used both pejoratively and in other contexts unrelated to hate speech \cite{Wang:2014:CET:2531602.2531734}. 

Syntactic features have been leveraged to better identify the targets and intensity of hate speech, for example sentences where a relevant noun and verb occur (e.g.\ \textit{kill} and \textit{Jews}) \cite{gitarieta15lexicon}, the POS trigram \enquote{DT jewish NN} \cite{Warner:2012:DHS:2390374.2390377}, and the syntactic structure I \textless intensity  \textgreater \ \textless user intent \textgreater \ \textless hate target \textgreater , e.g.\ \ \enquote{I f*cking hate white people} \cite{DBLP:conf/icwsm/SilvaMCBW16}.

Other supervised approaches to hate speech classification have unfortunately conflated hate speech with offensive language, making it difficult to ascertain the extent to which they are really identifying hate speech \cite{burnap2015cyber,DBLP:conf/naacl/WaseemH16}. Neural language models show promise in the task but existing work has used training data has a similarly broad definition of hate speech \cite{DBLP:conf/www/DjuricZMGRB15}. Non-linguistic features like the gender or ethnicity of the author can help improve hate speech classification but this information is often unavailable or unreliable on social media \cite{DBLP:conf/naacl/WaseemH16}.

\section{Data}
We begin with a hate speech lexicon containing words and phrases identified by internet users as hate speech, compiled by \textit{Hatebase.org}. Using the Twitter API we searched for tweets containing terms from the lexicon, resulting in a sample of tweets from 33,458 Twitter users. We extracted the time-line for each user, resulting in a set of 85.4 million tweets. From this corpus we then took a random sample of 25k tweets containing terms from the lexicon and had them manually coded by CrowdFlower (CF) workers. Workers were asked to label each tweet as one of three categories: hate speech, offensive but not hate speech, or neither offensive nor hate speech. They were provided with our definition along with a paragraph explaining it in further detail. Users were asked to think not just about the words appearing in a given tweet but about the context in which they were used. They were instructed that the presence of a particular word, however offensive, did not necessarily indicate a tweet is hate speech. Each tweet was coded by three or more people. The intercoder-agreement score provided by CF is 92\%. We use the majority decision for each tweet to assign a label. Some tweets were not assigned labels as there was no majority class. This results in a sample of 24,802 labeled tweets.

Only 5\% of tweets were coded as hate speech by the majority of coders and only 1.3\% were coded unanimously, demonstrating the imprecision of the Hatebase lexicon. This is much lower than a comparable study using Twitter, where 11.6\% of tweets were flagged as hate speech \cite{burnap2015cyber}, likely because we use a stricter criteria for hate speech. The majority of the tweets were considered to be offensive language (76\% at 2/3, 53\% at 3/3) and the remainder were considered to be non-offensive (16.6\% at 2/3, 11.8\% at 3/3). We then constructed features from these tweets and used them to train a classifier.

\section{Features}
We lowercased each tweet and stemmed it using the Porter stemmer,\footnote{We verified that the stemmer did not remove important information by reducing key terms to the same stem, e.g.\ \textit{f*gs} and \textit{f*ggots} stem to \textit{f*g} and \textit{f*ggot}.} then create bigram, unigram, and trigram features, each weighted by its TF-IDF. To capture information about the syntactic structure we use \texttt{NLTK} \cite{bird2009} to construct Penn Part-of-Speech (POS) tag unigrams, bigrams, and trigrams. To capture the quality of each tweet we use modified Flesch-Kincaid Grade Level and Flesch Reading Ease scores, where the number of sentences is fixed at one. We also use a sentiment lexicon designed for social media to assign sentiment scores to each tweet \cite{vader}. We also include binary and count indicators for hashtags, mentions, retweets, and URLs, as well as features for the number of characters, words, and syllables in each tweet.

\section{Model}
We first use a logistic regression with L1 regularization to reduce the dimensionality of the data. We then test a variety of models that have been used in prior work: logistic regression, na\"{i}ve Bayes, decision trees, random forests, and linear SVMs. We tested each model using 5-fold cross validation, holding out 10\% of the sample for evaluation to help prevent over-fitting. After using a grid-search to iterate over the models and parameters we find that the Logistic Regression and Linear SVM tended to perform significantly better than other models. We decided to use a logistic regression with L2 regularization for the final model as it more readily allows us to examine the predicted probabilities of class membership and has performed well in previous papers \cite{burnap2015cyber,DBLP:conf/naacl/WaseemH16}. We trained the final model using the entire dataset and used it to predict the label for each tweet. We use a one-versus-rest framework where a separate classifier is trained for each class and the class label with the highest predicted probability across all classifiers is assigned to each tweet. All modeling was performing using \texttt{scikit-learn} \cite{scikit-learn}. 

\begin{figure}
 \center
  \caption{\textbf{True versus predicted categories}}
  \includegraphics[width=0.5\textwidth]{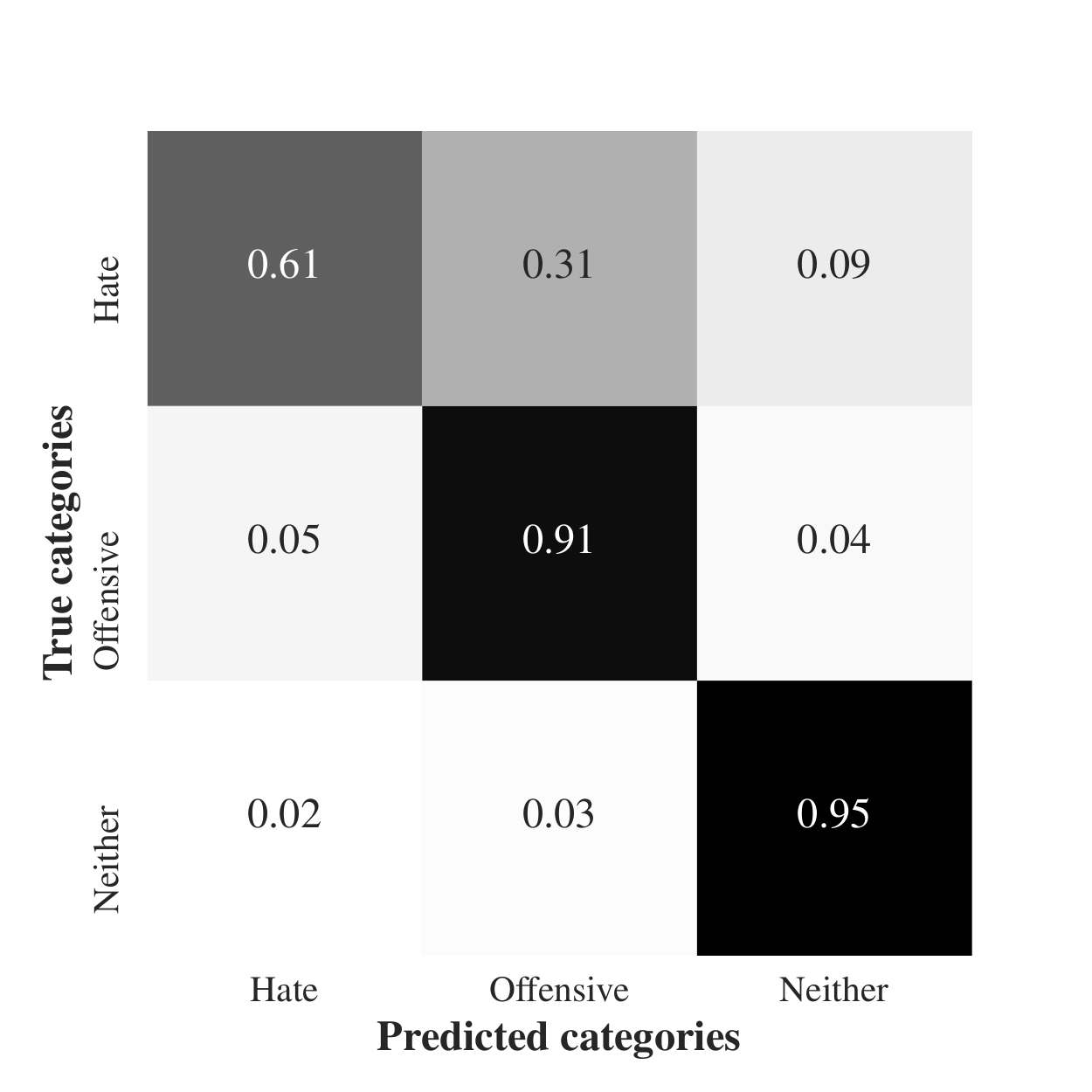}
\end{figure}

\section{Results}
The best performing model has an overall precision 0.91, recall of 0.90, and F1 score of 0.90. Looking at \textit{Figure 1}, however, we see that almost 40\% of hate speech is misclassified: the precision and recall scores for the hate class are 0.44 and 0.61 respectively. Most of the misclassification occurs in the upper triangle of this matrix, suggesting that the model is biased towards classifying tweets as less hateful or offensive than the human coders. Far fewer tweets are classified as more offensive or hateful than their true category; approximately 5\% of offensive and 2\% of innocuous tweets have been erroneously classified as hate speech. To explore why these tweets have been misclassified we now look more closely at the tweets and their predicted classes.

Tweets with the highest predicted probabilities of being hate speech tend to contain multiple racial or homophobic slurs, e.g. \textit{@JuanYeez shut yo beaner ass up sp*c and hop your f*ggot ass back across the border little n*gga} and \textit{RT @eBeZa: Stupid f*cking n*gger LeBron. You flipping jungle bunny monkey f*ggot}. Other tweets tend to be correctly identified as hate when they contained strongly racist or homophobic terms like \textit{n*gger} and \textit{f*ggot}. Interestingly, we also find cases where people use hate speech to respond to other hate speakers, such as this tweet where someone uses a homophobic slur to criticize someone else's racism: \textit{@MrMoonfrog @RacistNegro86 f*ck you, stupid ass coward b*tch f*ggot racist piece of sh*t}. 

Turning to true hate speech classified as offensive it appears that tweets with the highest predicted probability of being offensive are genuinely less hateful and were perhaps mislabeled, for example \textit{When you realize how curiosity is a b*tch \#CuriosityKilledMe} may have been erroneously coded as hate speech if people thought that \textit{curiosity} was a person, and \textit{Why no boycott of racist "redskins"? \#Redskins \#ChangeTheName} contains a slur but is actually against racism. It is likely that coders skimmed these tweets too quickly, picking out words or phrases that appeared to be hateful without considering the context. Turning to borderline cases, where the probability of being offensive is marginally higher than hate speech, it appears that the majority are hate speech, both directed towards other Twitter users, \textit{@MDreyfus @NatFascist88 Sh*t your ass your moms p*ssy u Jew b*stard. Ur times coming. Heil Hitler!} and general hateful statements like \textit{My advice of the day: If your a tranny...go f*ck your self!}. These tweets fit our definition of hate speech but were likely misclassified because they do not contain any of the terms most strongly associated with hate speech. Finally, the hateful tweets incorrectly labeled as neither tend not to contain hate or curse words, for example \textit{If some one isn't an Anglo-Saxon Protestant, they have no right to be alive in the US. None at all, they are foreign filth} contains a negative term, \textit{filth} but no slur against a particular group. We also see that rarer types of hate speech, for example this anti-Chinese statement \textit{Every slant in \#LA should be deported. Those scum have no right to be here. Chinatown should be bulldozed}, are incorrectly classified. While the classifier performs well at prevalent forms of hate speech, particularly anti-black racism and homophobia, but is less reliable at detecting types of hate speech that occur infrequently, a problem noted by \citeauthor{Nobata:2016:ALD:2872427.2883062} (\citeyear{Nobata:2016:ALD:2872427.2883062}).

A key flaw in much previous work is that offensive language is mislabeled as hate speech due to an overly broad definition. Our multi-class framework allows us to minimize these errors; only 5\% of our true offensive language was labeled as hate. The tweets correctly labeled as offensive tend to contain curse words and often sexist language, e.g. \textit{Why you worried bout that other h*e? Cuz that other h*e aint worried bout another h*e} and \textit{I knew Kendrick Lamar was onto something when he said \enquote{I call a b*tch a b*tch, a h*e a h*e, a woman a woman}}. Many of these tweets contain sexist terms like \textit{b*tch}, \textit{p*ssy}, and \textit{h*e}. Human coders appear to consider racists or homophobic terms to be hateful but consider words that are sexist and derogatory towards women to be only offensive, consistent prior findings \cite{DBLP:conf/naacl/WaseemH16}.

Looking at the tweets misclassified as hate speech we see that many contain multiple slurs, e.g. \textit{@SmogBaby: These h*es be lyin to all of us n*ggas} and \textit{My n*gga mister meaner just hope back in the b*tch}. While these tweets contain terms that can be considered racist and sexist it is apparent than many Twitter users use this type of language in their everyday communications. When they do contain racist language they tend to contain the term \textit{n*gga} rather than \textit{n*gger}, in line with the findings of \citeauthor{DBLP:conf/aaai/KwokW13} (\citeyear{DBLP:conf/aaai/KwokW13}). We  also found a few recurring phrases such as \textit{these h*es ain't loyal} that were actually lyrics from rap songs that users were quoting. Classification of such tweets as hate speech leads us to overestimate the prevalence of the phenomenon. While our model still misclassifies some offensive language as hate speech we are able to avoid the vast majority of these errors by differentiating between the two.

Finally, turning to the neither class, we see that tweets with the highest predicted probability of belonging to this class all appear to be innocuous and were included in the sample because they contained terms included in the Hatebase lexicon such as 
\textit{charlie} and \textit{bird} that are generally not used in a hateful manner. Tweets with overall positive sentiment and higher readability scores are more likely to belong to this class. The tweets in this category that have been misclassified as hate or offensive tend to mention race, sexuality, and other social categories that are targeted by hate speakers. Most appear to be misclassifications appear to be caused by on the presence of potentially offensive language, for example \textit{He's a damn good actor. As a gay man it's awesome to see an openly queer actor given the lead role for a major film} contains the potentially the offensive terms \textit{gay} and \textit{queer} but uses them in a positive sense. This problem has been encountered in previous research \cite{Warner:2012:DHS:2390374.2390377} and illustrates the importance of taking context into account. We also found a small number of cases where the coders appear to have missed hate speech that was correctly identified by our model, e.g. \textit{@mayormcgunn @SenFeinstein White people need those weapons to defend themselves from the subhuman trash your sort unleashes on us}. This finding is consistent with previous work that has found amateur coders to often be unreliable at identifying abusive content \cite{Nobata:2016:ALD:2872427.2883062,waseem:2016:NLPandCSS}.

\section{Conclusions}
If we conflate hate speech and offensive language then we erroneously consider many people to be hate speakers (errors in the lower triangle of Figure 1) and fail differentiate between commonplace offensive language and serious hate speech (errors in the upper triangle of Figure 1). Given the legal and moral implications of hate speech it is important that we are able to accurately distinguish between the two. Lexical methods are effective ways to identify potentially offensive terms but are inaccurate at identifying hate speech; only a small percentage of tweets flagged by the Hatebase lexicon were considered hate speech by human coders.\footnote{If a lexicon must be used we propose that a smaller lexicon with higher precision is preferable to a larger lexicon with higher recall. We have made a more restricted version of the Hatebase lexicon available here: \url{https://github.com/t-davidson/hate-speech-and-offensive-language}.} While automated classification methods can achieve relatively high accuracy at differentiating between these different classes, close analysis of the results shows that the presence or absence of particular offensive or hateful terms can both help and hinder accurate classification.

Consistent with previous work, we find that certain terms are particularly useful for distinguishing between hate speech and offensive language. While \textit{f*g}, \textit{b*tch}, and \textit{n*gga} are used in both hate speech and offensive language, the terms \textit{f*ggot} and \textit{n*gger} are generally associated with hate speech. Many of the tweets considered most hateful contain multiple racial and homophobic slurs. While this allows us to easily identify some of the more egregious instances of hate speech it means that we are more likely to misclassify hate speech if it doesn't contain any curse words or offensive terms. To more accurately classify such cases we should find sources of training data that are hateful without necessarily using particular keywords or offensive language.

Our results also illustrate how hate speech can be used in different ways: it can be directly send to a person or group of people targeted, it can be espoused to nobody in particular, and it can be used in conversation between people. Future work should distinguish between these different uses and look more closely at the social contexts and conversations in which hate speech occurs. We must also study more closely  the people who use hate speech, focusing both on their individual characteristics and motivations and on the social structures they are embedded in.

Hate speech is a difficult phenomenon to define and is not monolithic. Our classifications of hate speech tend to reflect our own subjective biases. People identify racist and homophobic slurs as hateful but tend to see sexist language as merely offensive. While our results show that people perform well at identifying some of the more egregious instances of hate speech, particularly anti-black racism and homophobia, it is important that we are cognizant of the social biases that enter into our algorithms and future work should aim to identify and correct these biases.

\fontsize{9.0pt}{10.0pt} \selectfont %uncomment to reduce min size allowed
\bibliographystyle{aaai}
\bibliography{hate_speech}

\begin{thebibliography}{}

\bibitem[\protect\citeauthoryear{Bird, Loper, and Klein}{2009}]{bird2009}
Bird, S.; Loper, E.; and Klein, E.
\newblock 2009.
\newblock {\em Natural Language Processing with Python}.
\newblock O'Reilly Media Inc.

\bibitem[\protect\citeauthoryear{Burnap and Williams}{2015}]{burnap2015cyber}
Burnap, P., and Williams, M.~L.
\newblock 2015.
\newblock Cyber hate speech on twitter: An application of machine
  classification and statistical modeling for policy and decision making.
\newblock {\em Policy \& Internet} 7(2):223--242.

\bibitem[\protect\citeauthoryear{Djuric \bgroup et al\mbox.\egroup
  }{2015}]{DBLP:conf/www/DjuricZMGRB15}
Djuric, N.; Zhou, J.; Morris, R.; Grbovic, M.; Radosavljevic, V.; and
  Bhamidipati, N.
\newblock 2015.
\newblock Hate speech detection with comment embeddings.
\newblock In {\em WWW},  29--30.

\bibitem[\protect\citeauthoryear{Gitari \bgroup et al\mbox.\egroup
  }{2015}]{gitarieta15lexicon}
Gitari, N.~D.; Zuping, Z.; Damien, H.; and Long, J.
\newblock 2015.
\newblock A lexicon-based approach for hate speech detection.
\newblock {\em International Journal of Multimedia and Ubiquitous Engineering}
  10:215--230.

\bibitem[\protect\citeauthoryear{Hutto and Gilbert}{2014}]{vader}
Hutto, C.~J., and Gilbert, E.
\newblock 2014.
\newblock {VADER:} {A} parsimonious rule-based model for sentiment analysis of
  social media text.
\newblock In {\em ICWSM}.

\bibitem[\protect\citeauthoryear{Jacobs and Potter}{2000}]{jacobs2000hate}
Jacobs, J.~B., and Potter, K.
\newblock 2000.
\newblock {\em Hate crimes: Criminal Law and Identity Politics}.
\newblock Oxford University Press.

\bibitem[\protect\citeauthoryear{Kwok and Wang}{2013}]{DBLP:conf/aaai/KwokW13}
Kwok, I., and Wang, Y.
\newblock 2013.
\newblock Locate the hate: Detecting tweets against blacks.
\newblock In {\em AAAI}.

\bibitem[\protect\citeauthoryear{Nobata \bgroup et al\mbox.\egroup
  }{2016}]{Nobata:2016:ALD:2872427.2883062}
Nobata, C.; Tetreault, J.; Thomas, A.; Mehdad, Y.; and Chang, Y.
\newblock 2016.
\newblock Abusive language detection in online user content.
\newblock In {\em WWW},  145--153.

\bibitem[\protect\citeauthoryear{Pedregosa and others}{2011}]{scikit-learn}
Pedregosa, F., et~al.
\newblock 2011.
\newblock Scikit-learn: Machine learning in {P}ython.
\newblock {\em Journal of Machine Learning Research} 12:2825--2830.

\bibitem[\protect\citeauthoryear{Silva \bgroup et al\mbox.\egroup
  }{2016}]{DBLP:conf/icwsm/SilvaMCBW16}
Silva, L.~A.; Mondal, M.; Correa, D.; Benevenuto, F.; and Weber, I.
\newblock 2016.
\newblock Analyzing the targets of hate in online social media.
\newblock In {\em ICWSM},  687--690.

\bibitem[\protect\citeauthoryear{Walker}{1994}]{walker1994hate}
Walker, S.
\newblock 1994.
\newblock {\em Hate Speech: The History of an American Controversy}.
\newblock U of Nebraska Press.

\bibitem[\protect\citeauthoryear{Wang \bgroup et al\mbox.\egroup
  }{2014}]{Wang:2014:CET:2531602.2531734}
Wang, W.; Chen, L.; Thirunarayan, K.; and Sheth, A.~P.
\newblock 2014.
\newblock Cursing in english on twitter.
\newblock In {\em CSCW},  415--425.

\bibitem[\protect\citeauthoryear{Warner and
  Hirschberg}{2012}]{Warner:2012:DHS:2390374.2390377}
Warner, W., and Hirschberg, J.
\newblock 2012.
\newblock Detecting hate speech on the world wide web.
\newblock In {\em LSM},  19--26.

\bibitem[\protect\citeauthoryear{Waseem and
  Hovy}{2016}]{DBLP:conf/naacl/WaseemH16}
Waseem, Z., and Hovy, D.
\newblock 2016.
\newblock Hateful symbols or hateful people? predictive features for hate
  speech detection on twitter.
\newblock In {\em SRW@HLT-NAACL},  88--93.

\bibitem[\protect\citeauthoryear{Waseem}{2016}]{waseem:2016:NLPandCSS}
Waseem, Z.
\newblock 2016.
\newblock Are you a racist or am i seeing things? annotator influence on hate
  speech detection on twitter.
\newblock In {\em Proceedings of the First Workshop on NLP and CSS},  138--142.

\end{thebibliography}

\end{document}